\documentclass{article} 

\usepackage{Apriel_Nemotron_15B}
\usepackage{comment}
\usepackage{multicol}

\usepackage[utf8]{inputenc} 
\usepackage[T1]{fontenc}    
\usepackage{newunicodechar}
\newunicodechar{ }{\,} 
\usepackage{hyperref}       
\usepackage{url}            
\usepackage{booktabs}       
\usepackage{amsfonts}       
\usepackage{nicefrac}       
\usepackage{microtype}      
\usepackage{xcolor}         
\usepackage{makecell}
\usepackage{tikz}
\usepackage{pgfplots}
\usepackage{graphicx}
\usepackage{subcaption}
\pgfplotsset{compat=1.18}
\pgfplotsset{scaled ticks=false}

\usepackage[
  backend=biber,
  style=numeric,
  sorting=none
]{biblatex}

\bibliography{Apriel_Nemotron_15B}

\title{Apriel-Nemotron-15B-Thinker}

\begin{document}

\maketitle
\begin{figure}[h]
    \centering
    \includegraphics[width=0.6\textwidth]{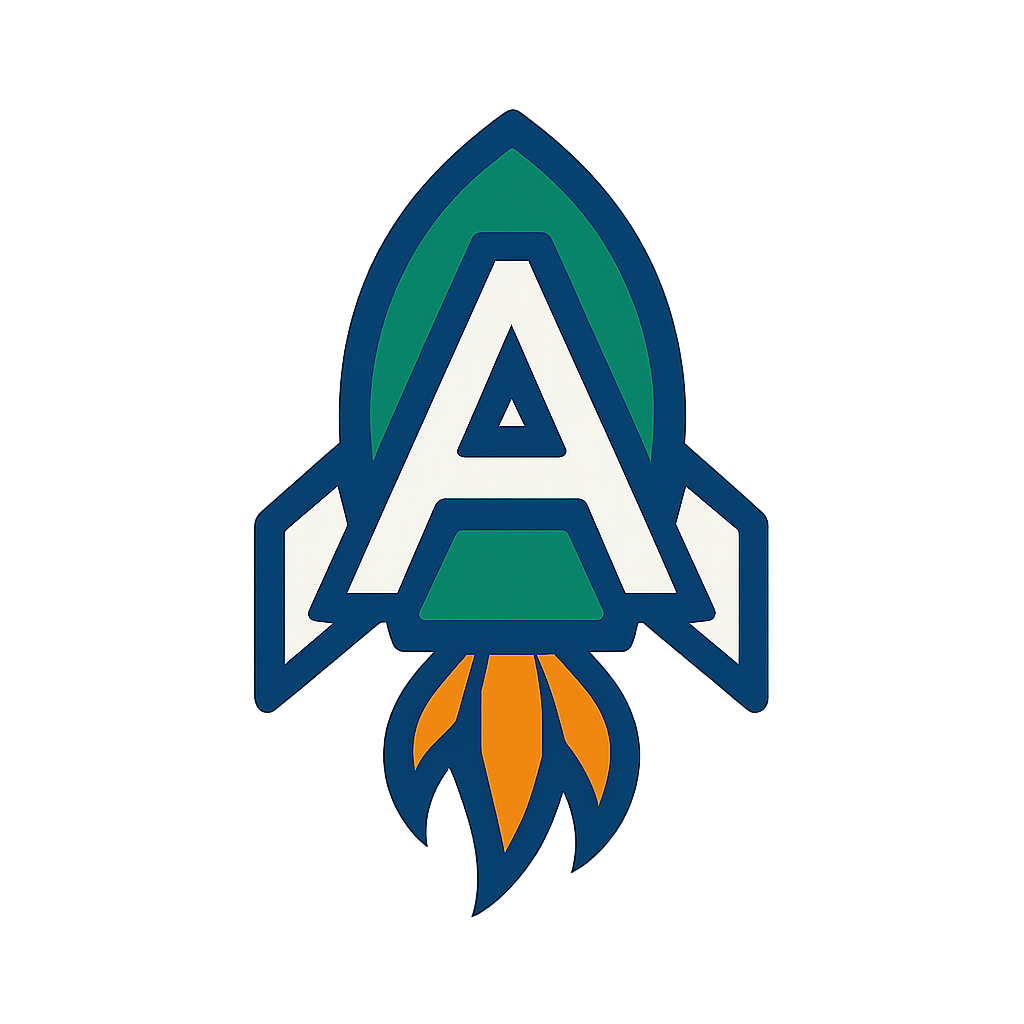}
\end{figure}

\begin{abstract}
  While large language models (LLMs) have achieved remarkable reasoning capabilities across domains like code, math and other enterprise  tasks, their significant \textbf{memory and computational costs} often preclude their use in practical enterprise settings. To this end, we introduce \textbf{Apriel-Nemotron-15B-Thinker}, a 15-billion parameter model in the ServiceNow Apriel SLM series that achieves performance against medium sized state-of-the-art models such as o1-mini, QWQ32B, and EXAONE-Deep-32B while maintaining only half the memory footprint of those alternatives. Apriel-Nemotron-15B-Thinker model is trained in a four stage training pipeline including 1) Base Model upscaling, 2) Continual Pre-training 3) Supervised Fine-tuning (SFT) and 4) Reinforcement Learning using GRPO. Comprehensive evaluations across a diverse suite of benchmarks consistently demonstrate that our Apriel-Nemotron-15B-Thinker model matches or exceeds the performance of its 32-billion parameter counterparts, despite being less than half their size.
\end{abstract}

\newpage
\section{Introduction}

Large Language Models (LLMs) have crossed the usability threshold—fueling breakthroughs in advanced reasoning, scientific discovery, and enterprise automation. With enough parameters,  effective pretraining, and specialized post-training, they can write code~\cite{roziere2023code, li2023starcoder}, explain complex concepts with step-by-step rationales~\cite{wei2022chain}, and solve Olympiad-level math problems {\cite{openai2025imo_x, deepmind2025gemini_imo}. However, achieving these capabilities has predominantly relied on increasingly large models (30B–180B parameters and beyond), whose memory footprint and inference cost keep them out of reach for many practical deployments. This is especially true for on-prem or air-gapped settings, common in enterprise software such as workflow orchestration and retrieval-augmented generation (RAG) pipelines.  On the other end of the spectrum, smaller “edge” models (2B–7B) offer impressive latency but continue to struggle with multi-step reasoning, tool invocation, and domain-specific tasks~\cite{abdin2024phi3technicalreporthighly, palm23, dumitru2024layer}.

These realities create a \emph{missing middle} i.e. models that are small enough to run on a single high-end GPU, yet smart enough to rival much larger baselines on multi-step reasoning, tool invocation, and domain-specific tasks. Catering to this gap would require development of models that (i) fit into 40 to 80 GB of memory, (ii) deliver optimal latency for real-time RAG or coding, and (iii) match the reasoning depth of much larger baselines.  Our work is specifically aimed at advancing this underexplored middle tier of model design.

We introduce Apriel-Nemotron-15B-\textsc{Thinker} \footnote{https://huggingface.co/ServiceNow-AI/Apriel-Nemotron-15b-Thinker}, a 15-billion parameter member of ServiceNow’s Apriel SLM family that is competitive to 30 to 32 billion parameter baselines such as O1-mini\footnote{Size of O1-Mini is not publicly known}~\cite{o1systemcard2024}, QWQ-32B~\cite{qwen2025qwq32b}, and EXAONE-Deep-32B~\cite{exaone2025deep32b} while fitting within the memory capacity of a single H100 or dual consumer GPUs

To reach this performance point we adopt a \textit{four‑stage} training pipeline that progressively sharpens reasoning while keeping compute under control.


 \textbf{(1) Model Upscaling} first expands a 12B backbone to 15B parameters through depth upscaling by duplicating transformer layers, and then trains the upscaled model on 100 billion tokens from a balanced open source mix, increasing the raw capability of the model.

 \textbf{(2) Continual Pre‑Training (CPT)} induces reasoning abilities in the model by training on around 70B tokens of text comprising of reasoning traces, chain-of-thought samples, and some pretraining text from the model upscaling phase. 

 \textbf{(3) Supervised Fine‑Tuning (SFT)} induces the reasoning behavior in which the model explicitly generates intermediate thought processes and reasoning steps prior to producing its final response. The model's reasoning capabilities are further enhanced in this phase comprising of about one million high-quality reasoning traces, encompassing tasks such as function calling, advanced mathematics, coding, science, retrieval-augmented generation (RAG) dialogues, etc. 

 \textbf{(4) Reinforcement Learning (RL)} uses Group Relative Policy Optimization (GRPO) with a rule-based reward system to strengthen robustness across diverse use cases, including output formatting, instruction following, advanced mathematics, coding, and agentic tool use. Rewards are based on structural adherence, correctness, and task success, using verifiable compositional instructions, executable code samples with test cases, and single-turn tool invocation scenarios. With 8-sample rollouts, KL regularization, and long-context generation enabled from the outset, this stage further refines the model’s reasoning precision, compliance, and reliability. 


We evaluate the resulting checkpoint on a comprehensive suite of benchmarks designed to reflect both practical deployment scenarios and evaluation benchmarks widely used in academic research. This includes seven enterprise-focused evaluations—\textit{MBPP}, \textit{BFCL-live-V2}, \textit{Enterprise-RAG}, \textit{MT-Bench}, \textit{MixEval}, \textit{IF-Eval}, and \textit{MultiChallenge}—as well as five reasoning-centric benchmarks: \textit{GPQA-Diamond}, \textit{MATH-500}, \textit{AIME-24}, \textit{AIME-25}, \textit{MMLU-Pro}, and \textit{AMC23}. Across this diverse evaluation landscape, \textit{Apriel-Nemotron-15B-Thinker} consistently demonstrates performance exceeding that of 30–32B models, while maintaining approximately half their memory footprint.

The remainder of this paper is organized as follows. \textbf{Section~2} presents the \textit{model upscaling} methodology, detailing parameter expansion, dataset composition, key hyperparameters, and preliminary results. \textbf{Section~3} describes the \textit{mid-training} pipeline, including Continual Pre-Training (CPT), Supervised Fine-Tuning (SFT), and accompanying ablation studies—covering domain-specialized SFT variants for function calling, retrieval-augmented generation (RAG), and advanced mathematics. \textbf{Section~4} outlines the \textit{post-training} phase, which consists of a lightweight alignment step using Group Relative Policy Optimization (GRPO). \textbf{Section~5} discusses the merging of SFT and GRPO checkpoints to obtain the final \textit{Apriel-Nemotron-15B-Thinker} model. Finally, \textbf{Section~6} presents comprehensive evaluation results on both enterprise-focused and academic benchmarks, with an analysis of token efficiency.

Our key contributions are summarized as follows:

\begin{itemize}
    \item A compute-efficient depth-scaling strategy that transforms a 12B model into a stable 15B model using less than $20\%$ of the compute required for training from scratch.
    
    \item Empirical study and evidence that a reasoning-centric CPT curriculum improves performance on mathematics and coding tasks by up to $20\%$ without the use of task-specific labels.
    
    \item A \emph{specialize–merge} SFT strategy that yields simultaneous improvements of more than $7\%$ on AIME-24 and more than $5\%$ on BFCL-Live.
    
    \item A lightweight GRPO-based alignment framework that enhances robustness on enterprise RAG, function calling, and MT-Bench tasks, while remaining within typical on-prem hardware constraints.
    
    \item State-of-the-art performance across seven enterprise benchmarks and five academic reasoning tasks, achieved with approximately half the memory footprint of leading 30B parameter models.
\end{itemize}

\section{Model Upscaling}
Recent work on emergent behaviors \cite{kaplan2020scalinglawsneurallanguage, berti2025emergentabilitieslargelanguage} in LLMs posits that model size is a crucial factor determining sudden emergent capabilities. Thus our initial investigation looked into what minimal model size is capable of advanced reasoning required for the complex enterprise tasks in our scope. We empirically find that the lower bound of the number of parameters of such a model is 15B parameters. 

We start with the Mistral-Nemo-Base-2407 \footnote{https://huggingface.co/mistralai/Mistral-Nemo-Base-2407} base model (12B parameters) and find it lacking for balancing performance on complex reasoning capabilities for math, coding and other benchmarks. We iteratively ablate the model size by increasing parameters until we reached a sweet spot of 15B parameters balancing compute, latency and performance while still being deployable on a single high end GPU.


\subsection{Increasing Parameters}
Inspired by approaches in \cite{kim2024solar107bscalinglarge} and \cite{Falcon3}, we experiment with two strategies to increase the model parameters, i.e., width-upscaling and depth-upscaling. For width-upscaling we introduce additional parameters along the width dimension by making the MLP intermediate dimension wider. For depth-upscaling, we create additional transformer layers using various approaches including averaging, max-pooling, averaging alternative layers, and duplicating layers. We observe that duplication of intermediate layers lets model starts with the lowest training loss and prefer it for training stability. Width-upscaling, on the other hand, started out highly unstable, so we defer this experimentation to future work. 


\subsection{Dataset}
Upscaling model parameters yields lower results without fine-tuning, and hence we see the need to continually pretrain the model. Since the training dataset for Mistral-Nemo-Base-2407 is not released, we use an existing open-source corpus 
as a proxy for the replay data. We use 50\% of this along with dataset sourced from diverse domains, including high-quality web content, scientific, technical literature, reference works, programming code, and mathematical problem sets, stackexchange, coding data from different programming languages, etc. We train the model on this mix for ~100B tokens.

The model was trained on 100B tokens with a learning rate of 5e-5, batch size of 768, and sequence length of 16k tokens, yielding 12 million tokens per batch.

We also experiment with creating smaller models (10B, 8B) by dropping layers from the 12B parameter model. However, reasoning models trained on those sizes do not yield comparative results immediately, and hence we conclude that the model scale matters significantly for reasoning models. Model up-scaling also provides us with a way to create models of desired size and latency from a given pretrained, open-source model. This is a relatively compute-efficient model which avoids pertaining to models of various sizes from scratch.


\subsection{Pretraining Results}

\begin{figure}[htb]
    \centering
    \begin{subfigure}{0.325\textwidth}
        \centering
        \includegraphics[width=\textwidth]{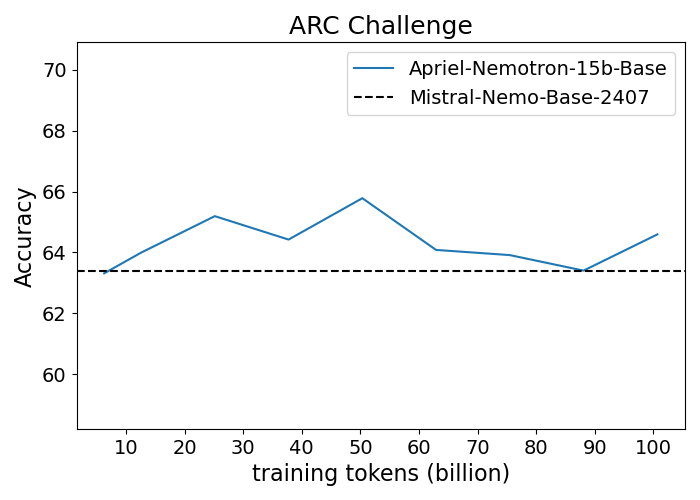}
    \end{subfigure}
    \hfill
    \begin{subfigure}{0.325\textwidth}
        \centering
        \includegraphics[width=\textwidth]{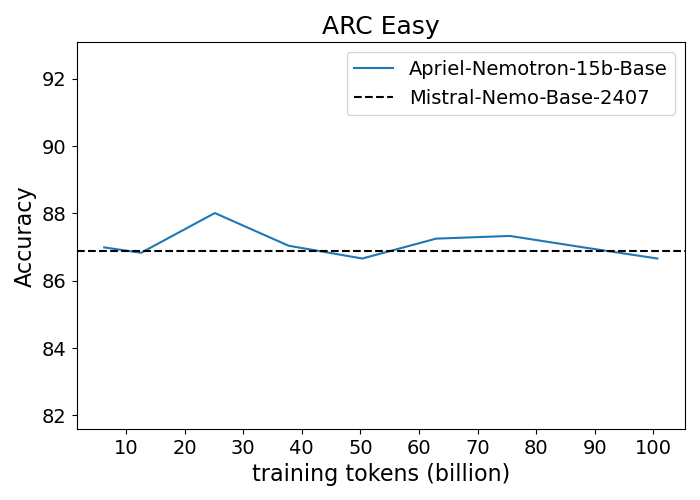}
    \end{subfigure}
    \hfill
    \begin{subfigure}{0.325\textwidth}
        \centering
        \includegraphics[width=\textwidth]{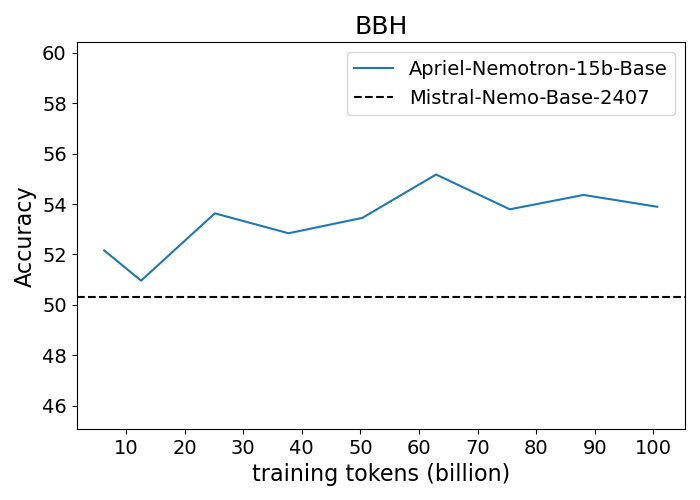}
    \end{subfigure}
    
    \vskip\baselineskip
    
    \begin{subfigure}{0.325\textwidth}
        \centering
        \includegraphics[width=\textwidth]{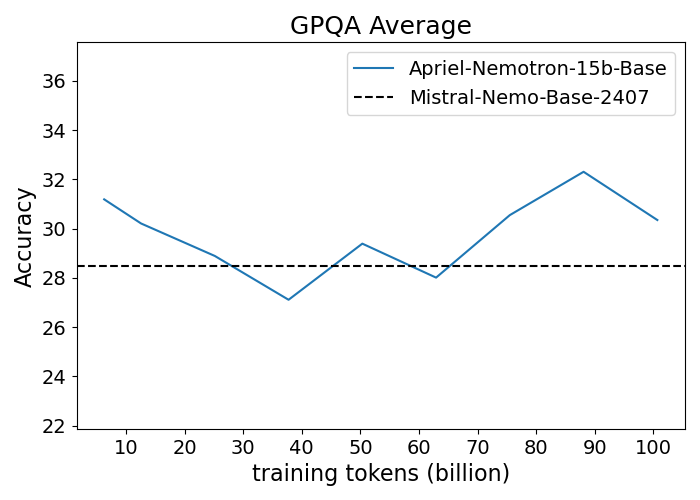}
    \end{subfigure}
    \hfill
    \begin{subfigure}{0.325\textwidth}
        \centering
        \includegraphics[width=\textwidth]{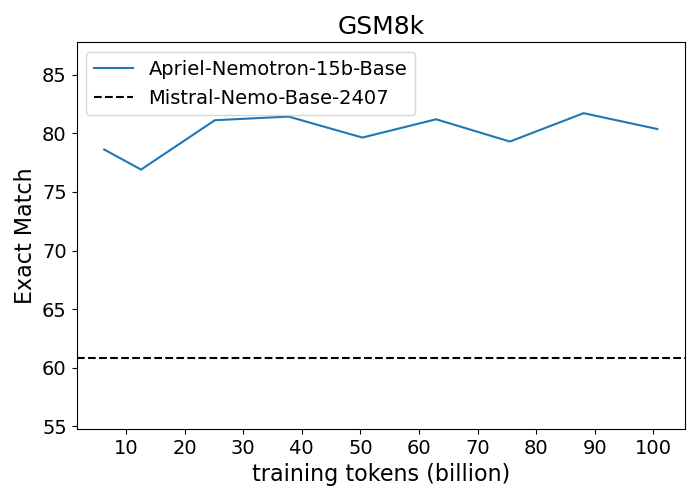}
    \end{subfigure}
    \hfill
    \begin{subfigure}{0.325\textwidth}
        \centering
        \includegraphics[width=\textwidth]{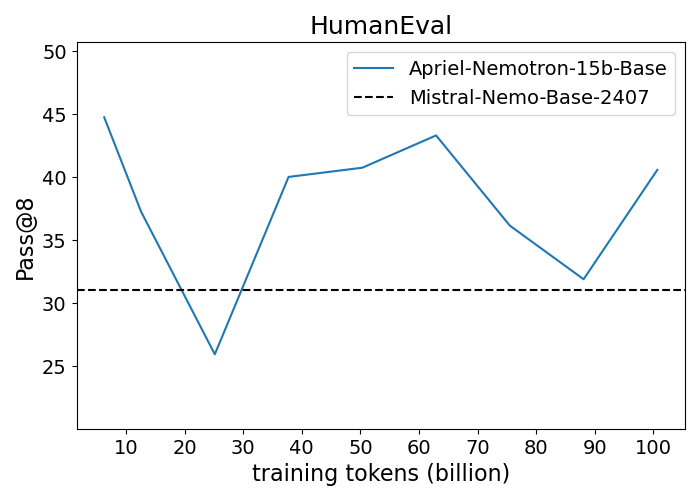}
    \end{subfigure}

    \vskip\baselineskip
    
    \begin{subfigure}{0.325\textwidth}
        \centering
        \includegraphics[width=\textwidth]{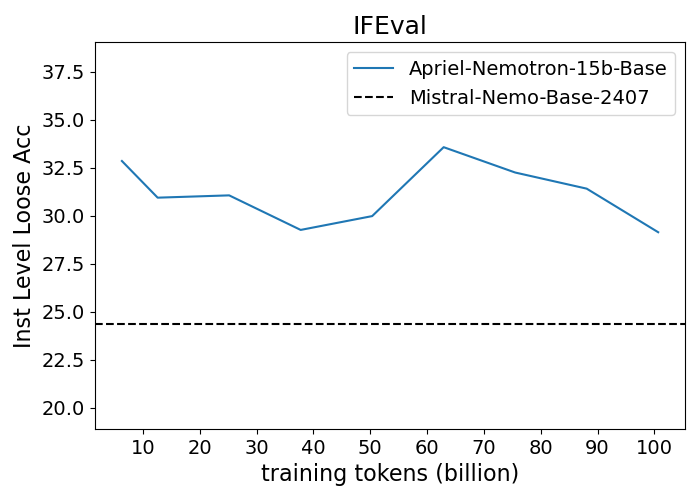}
    \end{subfigure}
    \hfill
    \begin{subfigure}{0.325\textwidth}
        \centering
        \includegraphics[width=\textwidth]{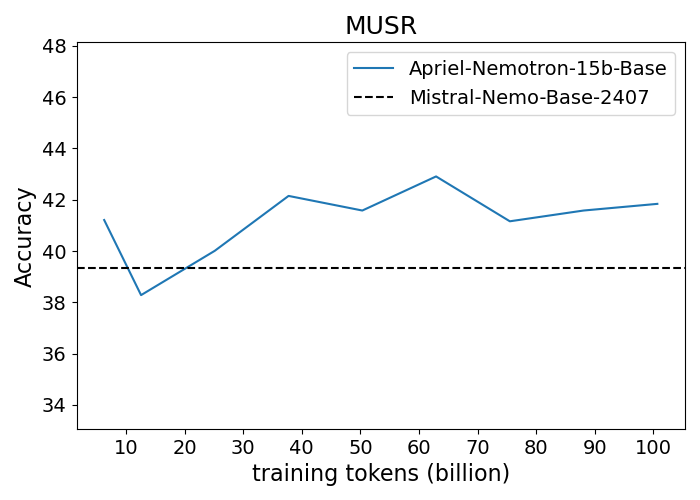}
    \end{subfigure}
    \hfill
    \begin{subfigure}{0.325\textwidth}
        \centering
        \includegraphics[width=\textwidth]{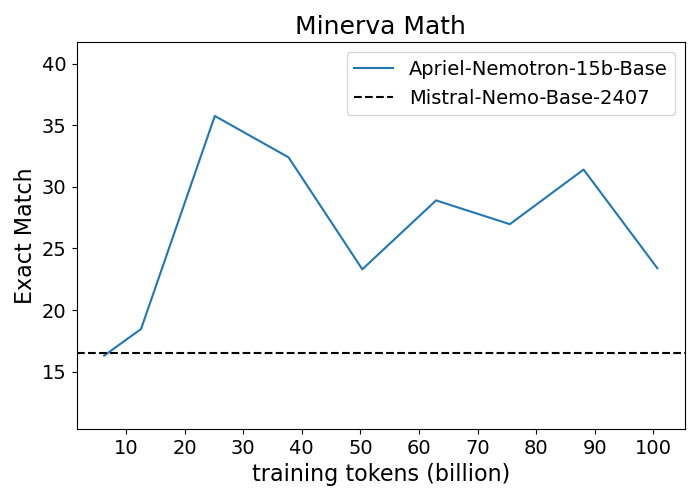}
    \end{subfigure}

    \vskip\baselineskip
    
    \begin{subfigure}{0.325\textwidth}
        \centering
        \includegraphics[width=\textwidth]{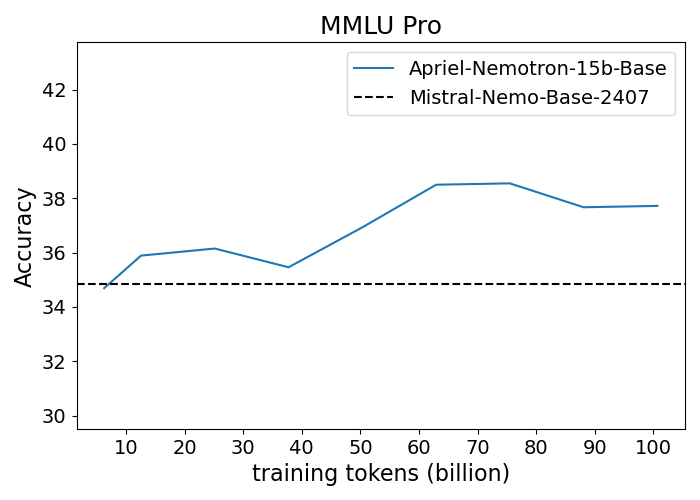}
    \end{subfigure}
    \hfill
    \begin{subfigure}{0.325\textwidth}
        \centering
        \includegraphics[width=\textwidth]{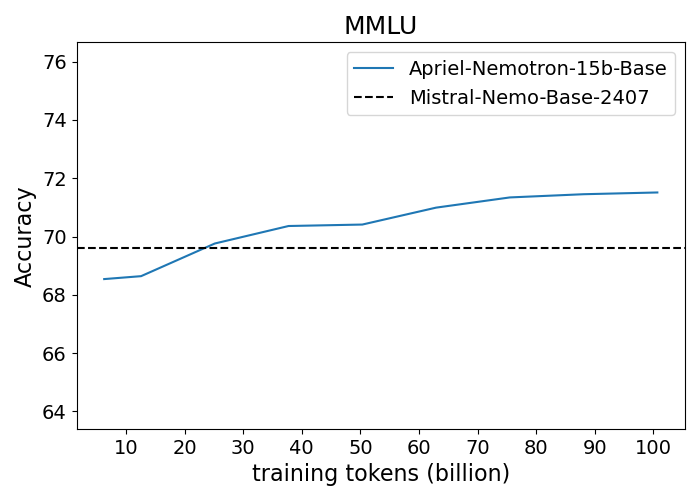}
    \end{subfigure}
    \hfill
    \begin{subfigure}{0.325\textwidth}
        \centering
        \includegraphics[width=\textwidth]{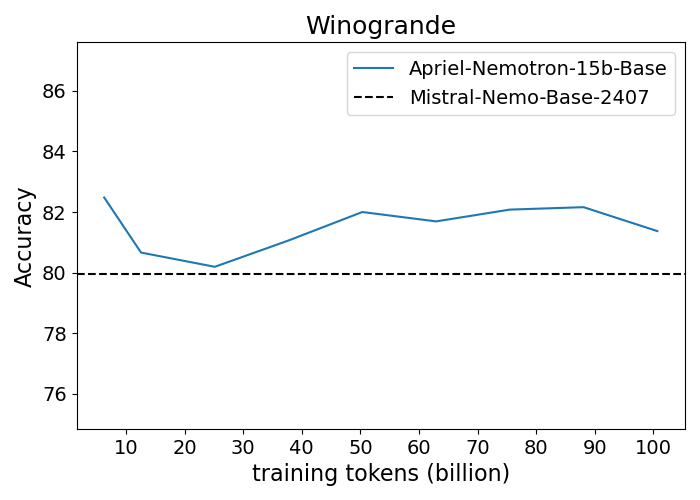}
    \end{subfigure}
    
    \caption{Figure showing the performance progression of Apriel-Nemotron-15B-Base and the static baseline performance of Mistral-Nemo-Base-2407 (represented by a straight dotted line) across twelve downstream benchmarks. The x-axis indicates cumulative pretraining tokens (in billions, from 10B to 100B), and the y-axis shows the relevant performance metric (e.g., Accuracy, Exact Match). Benchmarks include ARC Challenge, ARC Easy, BBH, GPQA Average, GSM8K, HumanEval, IfEval, MUSR, Minerva Math, MMLU Pro, MMLU, and WinoGrande, providing a comprehensive view of model capabilities in reasoning, code generation, mathematical problem-solving, and general language understanding.
    }
    \label{fig:main}
\end{figure}

We evaluate the impact of training data scale on model performance across 12 diverse benchmarks. Figure \ref{fig:main} shows the performance on these benchmarks as training progresses up to 100 billion tokens, with Mistral-Nemo-Base-2407 as the baseline.

The benchmarks exhibited three distinct patterns of improvement as training data scaled from 10B to 100B tokens. The first group, comprising mathematical reasoning, code generation, and knowledge-intensive tasks (GSM8K, HumanEval, BBH, and MMLU), showed the most substantial gains. GSM8K improved from 75\% to 85\% accuracy, MMLU rose from 68\% to 72\%, while HumanEval increased from 30\% to 42\% despite significant volatility during training. BBH demonstrated steady growth from 52\% to 56\%, with all benchmarks in this group substantially outperforming their respective Mistral baselines.

A second group of benchmarks showed moderate improvements with relatively steady gains throughout training. This included reasoning tasks (ARC Challenge and ARC Easy), knowledge evaluations (IFEval and MUSR), and MMLU Pro. These benchmarks typically gained 2-4 percentage points over the training range, with improvements from 64\% to 66\% on ARC Challenge and 27.5\% to 32\% on IFEval. The gains in this group were more gradual and consistent compared to the first group.

The third category consisted of benchmarks that showed minimal improvement or high variability throughout training. WinoGrande improved only marginally from 81\% to 82.5\%, while GPQA Average gained just 2 percentage points despite starting from a low baseline of 28\%. Minerva Math displayed particularly erratic behavior, fluctuating between 25\% and 35\% throughout training and ultimately ending at 27\%, slightly below its starting point.

The improvement patterns varied significantly across benchmarks, with some showing continuous gains throughout training (GSM8K, MMLU), others exhibiting early saturation (WinoGrande, ARC tasks), and several displaying non-monotonic behavior with fluctuations at different training stages (HumanEval, Minerva Math). Overall, the Apriel-Nemoxtron-15b-Base model consistently outperformed the Mistral-Nemo-Base-2407 baseline across 11 of the 12 benchmarks, demonstrating the effectiveness of increased training data scale for improving model capabilities.

\section{Mid Training}
After scaling up our model, we enhanced its reasoning abilities during the midtraining phase. The midtraining phase consisted of two stages - continual pretraining (CPT) and supervised fine-tuning (SFT) - with model merging applied in both stages.

\subsection{Continual Pretraining (CPT)}
To enhance the reasoning capabilities of the base model, we continually pretrained the model using a diverse corpus that included reasoning samples across domains like math, science, coding, and instruction-following. This was supplemented with Chain-of-Thought (CoT) data in the same domains and generic pretraining-style content as replay from the upscaling stage, ensuring the model maintains its foundational capabilities while developing improved reasoning abilities. The training mixture consisted of 60\% reasoning samples, 25\% CoT samples, and 15\% pretraining-style samples.

During continual pretraining, no chat template elements were used. For reasoning samples, the complete sequence of input, intermediate reasoning steps, and target outputs were concatenated into a single string with newline separators. CoT samples maintained the same newline-delimited format with input-output pairs. Loss was computed on all tokens throughout the sequence. The training was run on 68B tokens with a batch size of 768. Samples were packed to sequence length of 16k, but cross-document attention was prevented by masking. We used the AdamW optimizer for training with weight decay of 0.1 and a base learning rate of 5e-5, cosine-decayed to 5e-6, with linear warmup of 10\% of the total steps. Following \cite{openmathinstruct2}, we average 3 equally spaced checkpoints from the CPT run to prepare the final model for the next stage of training. 

We evaluate CPT effectiveness through two approaches: (1) performance on diverse benchmarks from the Language Model Evaluation Harness \cite{eval-harness}, and (2) downstream reasoning performance after supervised fine-tuning with a small SFT set containing 15k reasoning samples, which was run for 3 epochs with a batch size of 128 samples at a maximum sequence length of 16k, a learning rate of 1e-5 cosine decayed to 0 (with 10\% linear warmup) and no sequence packing.
Table~\ref{tab:model_comparison} shows that at the CPT stage, we observe improvements in math, science, and instruction-following benchmarks, while commonsense reasoning performance dips slightly. Table~\ref{tab:sft_cpt_comparison} demonstrates that after SFT, reasoning benchmarks show substantial performance gains across tasks.

\begin{table}[ht]
\centering
\begin{tabular}{lccccccccccccc}
\toprule
\textbf{Benchmark} & \textbf{Before CPT} & \textbf{After CPT} \\
\midrule
Arc            & 68.51 & 65.27 \\
GSM8k          & 78.54 & 82.03 \\
Hellaswag      & 85.32 & 83.46 \\
BBH            & 57.52 & 57.17 \\
GPQA           & 31.63 & 32.71 \\
MMLU Pro       & 40.57 & 41.60 \\
MUSR           & 41.79 & 40.60 \\
MBPP           & 66.60 & 63.20 \\
TruthfulQA     & 53.39 & 51.38 \\
Winogrande     & 80.97 & 79.71 \\
Minerva Math   & 45.30 & 49.20 \\
IF Eval        & 22.36 & 36.78 \\
MMLU           & 72.82 & 71.87 \\
\midrule
\textbf{Average} & 57.33 & 58.08 \\
\bottomrule
\end{tabular}
\vspace{+2mm}
\caption{Benchmark performance comparison before and after continual pretraining (CPT)}
\label{tab:model_comparison}
\end{table}

\begin{table}[ht]
\centering
\begin{tabular}{lcc}
\toprule
\textbf{Benchmark} & \textbf{SFT on model before CPT} & \textbf{SFT on model after CPT} \\
\midrule
GPQA Diamond & 37.20 & 46.46 \\
MATH-500     & 80.40 & 90.80 \\
AIME'24       & 16.00 & 58.00 \\
AIME'25     & 18.44 & 45.99 \\
AMC23     & 59.50 & 96.00 \\
\bottomrule
\end{tabular}
\vspace{+2mm}
\caption{Reasoning benchmark performance after supervised fine-tuning (SFT) on a 15B model before and after continual pretraining (CPT). The SFT was performed on a small SFT set containing 15k reasoning samples and evaluated on benchmarks relevant to the dataset.}
\label{tab:sft_cpt_comparison}
\end{table}

\subsection{Supervised Fine Tuning (SFT)}
Following the upscaling and CPT stages, which yielded a base model with strong reasoning capabilities, we performed Supervised Fine-Tuning (SFT) to develop it into a full-fledged reasoner. Similar to observations in \cite{nemotron4340btechnicalreport}, we found that attempting to learn multiple tasks concurrently had some negative interference that prevented optimal performance across all the tasks being learned. However, we also observed some beneficial cross-domain transfer effects \cite{maheshwary2025m2lingual}, where training on one domain (such as math) improved performance on other domains (such as coding). To balance between negative interference and positive transfer, we trained specialized models maintaining a common core of training data while varying the domain-specific portions. We then merged the models using by averaging the weights of the specialized models in specified proportions.

This approach also helped us address the uneven availability of reasoning samples across domains. For instance, we had higher proportion of samples in domains like math and coding than in domains like function calling and other enterprise applications. Training specialized models allowed us to maximally utilize all available data within each domain while preventing data imbalances from adversely affecting performance in any particular area. The common core of shared data ensured consistent foundational capabilities, while domain-specific portions allowed for specialized expertise. 

All training runs were run with learning rate of 1e-5 cosine decayed to 0 over all the epochs, with a linear warmup of 10\% of the first epoch. While standard SFTs were run for 3 or 4 epochs, some specialized SFTs were run up to 7 epochs as described in subsequent sections. Sequences were packed to a length of 32K, and loss was computed only on the assistant turns.

Below, we note some of our observations from training the specialized models.

\subsubsection{Function Calling and RAG}
For tasks like function calling and RAG that are directly related to enterprise applications, we observed that a balanced SFT data mix from all domains yielded the best results. This mix contained a slight over-representation of data from instruction following, function calling, RAG, coding, and multi-turn conversations, comprising approximately 1 million samples in total, and the training was run for 3 epochs.

\subsubsection{Advanced Math}
To improve performance on benchmarks measuring advanced/olympiad-level mathematical abilities (such as AIME24 and AIME25), we observed benefits from training for a large number of steps with data containing multiple responses for a single prompt. Having multiple responses per prompt likely aids in preventing overfitting when training for a large number of steps. Significant improvements were observed when training on a dataset of approximately 200K samples with an over-representation of math data for 8 epochs. For the math samples, we enforced that each prompt had at least 3-4 generations. These observations align with findings reported in \cite{openmathreasoning}.


\begin{figure}[htp]
    \centering
    \includegraphics[width=10cm]{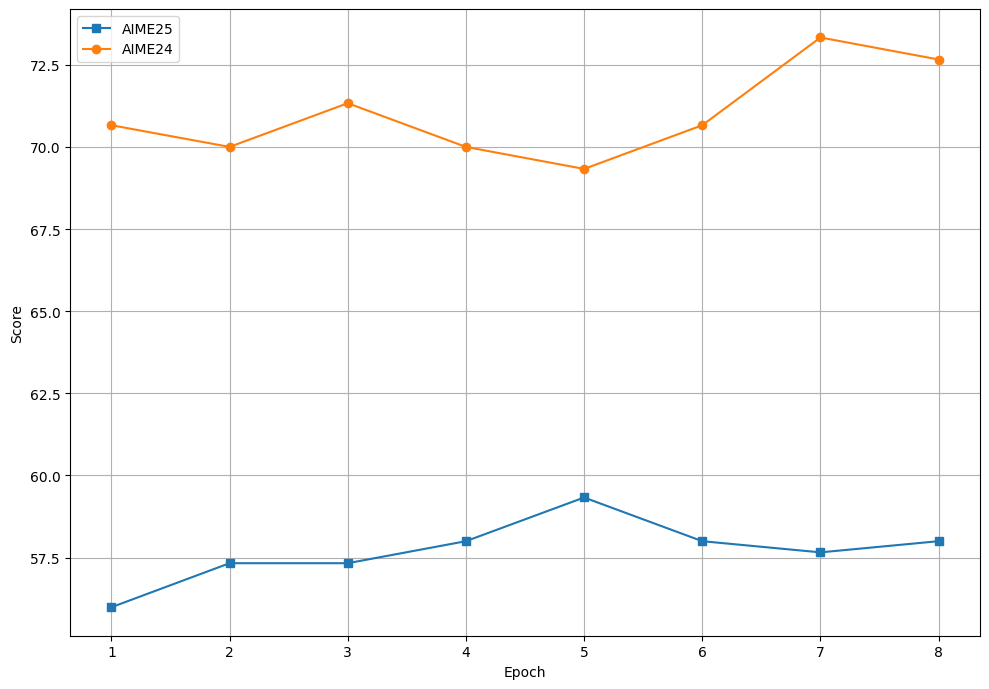}
    \caption{Variation of AIME24 and AIME24 scores over 8 epochs of SFT with a dataset of size 200K. We notice an overall upwards trend with some fluctuations between epochs.}
    \label{fig:galaxy}
\end{figure}

\subsubsection{Other Benchmarks}

On benchmarks like MATH500, which contains high-school level competition mathematics problems, we observed consistent performance across multiple checkpoints comparable to our best models, suggesting that the model had already developed sufficient mathematical reasoning capabilities during the CPT stage. To validate the effectiveness of the base model on these benchmarks, we conducted a much smaller SFT with around 15k reasoning samples which also performed similarly to the larger SFTs on MATH500, but scored lower on more advanced mathematics benchmarks like AIME24 and AIME25.

We observed similar trends on other benchmarks such as MBPP (Python coding), IF-Eval (instruction following), and MT-Bench. However, these capabilities were further enhanced during the post-training stages described in subsequent sections.

\section{Post training}

\subsection{RL}

The RL phase is designed to significantly improve the overall performance and robustness of the model in a wide range of use cases. We developed an advanced rule-based reward system for the following use cases:
\begin{itemize}
    \item \textbf{Output Format:} We expect the model to include both its reasoning process and the final response, each enclosed within predefined tags. For every prompt, we first verify that the output conforms to this expected tag structure. 

    \item \textbf{Advanced Math:} To improve the model’s mathematical reasoning capabilities, we generated eight candidate solutions for each of approximately 100,000 prompts using the SFT model. We then curated a targeted subset of 18,000 prompts for which the model produced at least one correct solution and at least three incorrect ones. This subset was used to train the model with a focus on improving its ability to distinguish correct reasoning paths from common failure modes.

    \item \textbf{Instruction Following:} We employed verifiable compositional instructions (~14,000) that constrain responses in terms of content, format, length, and structure. This significantly improves the model’s ability to interpret and adhere to user directives.

    \item \textbf{Coding:} We included 30,000 verifiable Python and JavaScript code samples, each accompanied by multiple test cases. The reward is determined by the percentage of test cases in which the model output passes.

    \item \textbf{Agentic Ability:} The RL training stage also includes approximately 32,000 single-turn agentic scenarios that require the invocation of one or more tools.

\end{itemize}

Since the SFT model is already capable of generating long reasoning traces, the maximum generation length is fixed at 32,768 tokens from the beginning of the GRPO stage \cite{shao2024deepseekmathpushinglimitsmathematical}. We used a sampling temperature of 1.0 and a top-p value of 0.95 to generate 8 samples from each prompt. The RL training hyperparameters are as follows: a batch size of 512 during rollout and training, batch size was 512 and we employed a learning rate of $1 \times 10^{-6}$ and a KL regularization of 0.001.



\section{The Final Model}
\begin{figure}[htp]
    \centering
    \includegraphics[width=12.5cm]{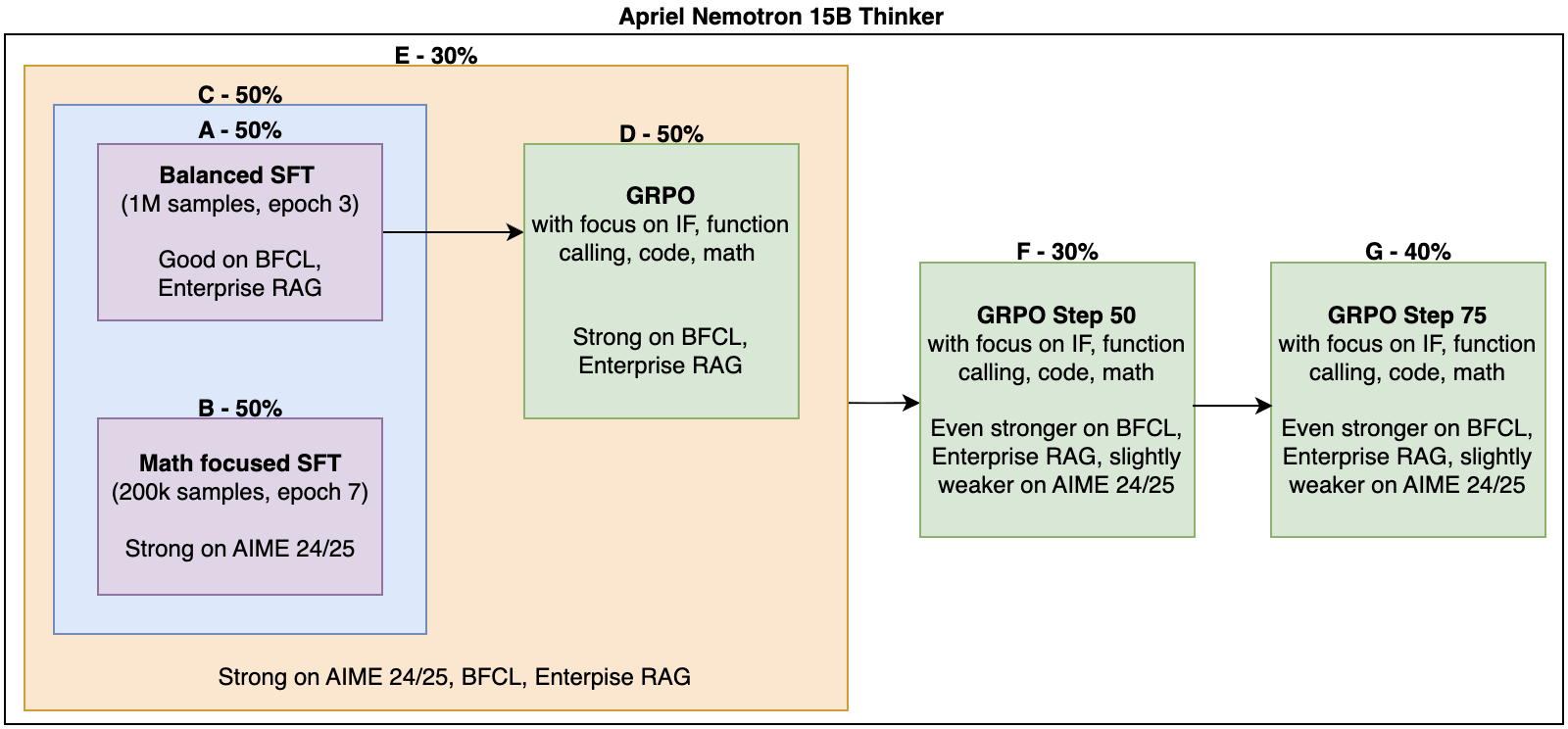}
    \caption{\textbf{Construction of the final Apriel Nemotron 15B Thinker checkpoint from the SFT and GRPO stages} Each labeled box (A–G) represents a distinct model checkpoint. Groups of boxes enclosed within a larger box indicate that the corresponding checkpoints were merged, with the percentages above each box denoting their contribution to the resulting merged checkpoint. For example, the final Apriel Nemotron 15B model was obtained by merging checkpoints E, F, and G in proportions of 30\%, 30\%, and 40\%, respectively. Arrows denote continued training from one checkpoint to the next.}
    \label{fig:final_model}
\end{figure}

To construct Apriel Nemotron 15B Thinker, we employed a staged training and merging strategy that integrates both SFT and GRPO. The training runs and merges are detailed in Figure \ref{fig:final_model}. Checkpoints A and B correspond to SFT-trained models: a balanced SFT model (A), trained on 1M samples for 3 epochs, exhibiting strong performance on BFCL and Enterprise RAG; and a math-focused SFT model (B), trained on 200k samples for 8 epochs (we select the 7th epoch for the final model), which performs well on AIME 24/25. These were merged in equal proportion to form checkpoint C. Separately, checkpoint D was obtained via GRPO on top of checkpoint A and then merged in equal proportions with C to produce checkpoint E. Continued GRPO training on top of E yielded checkpoints F and G, which further improved performance on BFCL and Enterprise RAG while introducing a slight regression on AIME 24/25. The final model was obtained by merging checkpoints E, F, and G in proportions of 30\%, 30\%, and 40\%, respectively, to balance the gains from the GRPO run with the slightly stronger math performance of the base SFT checkpoint. We used \verb|mergekit| \cite{goddard-etal-2024-arcees} to merge model checkpoints.

\section{Benchmarking}
\label{Benchmarking}
To evaluate \textsc{Apriel-Nemotron}-15B-\textsc{Thinker}, we selected benchmarks targeting capabilities central to enterprise applications, including function calling, instruction following, and complex multi-step problem solving, as well as tasks that assess interdependent skills in logical reasoning, mathematical problem solving, and code generation. Additionally, we evaluate all models on our internal enterprise RAG benchmark.

All evaluations were conducted under zero-shot conditions using a sampling temperature of 0.6 and a maximum generation length of 32,000 tokens.

\subsection{Enterprise Benchmarks}
To assess each model’s suitability for real-world enterprise use cases, we evaluate them across seven benchmarks covering code generation (MBPP\cite{austin2021program}), function calling (BFCL-live-V2\cite{evalchemy2023}), retrieval-augmented QA (Enterprise RAG), conversational flow and instruction-following (MT Bench), broad spectrum capabilities (MixEval\cite{ni2024mixevalderivingwisdomcrowd}), instruction following (IFEval\cite{zhou2023instructionfollowingevaluationlargelanguage}) and complex multi‐step problem solving (MultiChallenge\cite{sirdeshmukh2025multichallengerealisticmultiturnconversation}). These tasks collectively stress the core capabilities required in production AI systems—correctness, robustness, domain‐adaptivity, and sustained reasoning—under zero‐shot conditions. Table \ref{tab:benchmark-results} reports each model’s performance, where higher scores indicate stronger enterprise readiness.

\begin{table}[ht]
  \centering
  \small
  \begin{tabular}{@{} l c c c c c c @{}}
    \toprule
    \textbf{Benchmark} & 
    \textbf{\makecell{RekaAI\\Flash-3}} & 
    \textbf{\makecell{Nvidia\\Nemotron\\Nano-8B}} & 
    \textbf{\makecell{Qwen\\QWQ-32B}} & 
    \textbf{\makecell{OpenAI\\O1-mini}} & 
    \textbf{\makecell{LGAI\\ExaOne-32B}} & 
    \textbf{\makecell{Apriel\\Nemotron-15B}} \\
    \midrule
    MBPP (0-shot, pass@1)        & 80.2 & 73.8 & 88.2 & 93.1 & 76.8 & 85.8 \\
    BFCL-live-V2 & 77.37 & 54.15 & 79.0 & 81.03 & 75.42 & 75.43 \\
    Enterprise RAG & 57.9 & 11.1 & 65.2 & 66.5 & 52.1 & 69.2 \\
    MT Bench & 8.431 & 7.425 & 8.456 & 8.381 & 8.394 & 8.569 \\
    MixEval & 79.1 & 62.1 & 77.34 & 82.9 & 80.6 & 82.79 \\
    IFEval (strict-prompt) & 81.6 & 69.78 & 82.8 & 79.5 & 83.1 & 84.6 \\
    MultiChallenge & 24.5 & 16.11 & 37.7 & 30.8 & 38.5 & 36.6 \\
    \bottomrule
  \end{tabular}
  \vspace{4mm}
  \caption{Performance comparison of various LLMs across enterprise-oriented benchmarks, including code generation (MBPP), functional correctness (BFCL-live-V2), enterprise retrieval-augmented generation (Enterprise RAG), multi-turn conversation quality (MT Bench), mixed-domain reasoning (MixEval), instruction-following (IFEval), and complex multi-task reasoning (MultiChallenge).}
  \label{tab:benchmark-results}
\end{table}

Despite its mid-sized footprint, Apriel-Nemotron-15b-thinker delivers top-tier performance—leading on MT Bench and IFEval, placing second on mixed-domain (MixEval) and code generation (MBPP), and holding its own in reasoning (BFCL-live) and RAG.

\subsection{Academic Reasoning Benchmarks}
To assess various facets of "academic intelligence" on \textsc{Apriel-Nemotron}-15B-\textsc{Thinker} and similar models, we selected several benchmarks to evaluate expert-level reasoning, broad knowledge recall, and the synthesis of executable code. These benchmarks evaluate advanced mathematical and logical reasoning (AIME, MATH-500), expert-level scientific knowledge and reasoning (GPQA-Diamond), broad domain-general understanding (MMLU-Pro \cite{wang2024mmluprorobustchallengingmultitask}) and executable code synthesis and programmatic knowledge (LiveCodeBench). 

All evaluations use standardized prompts and scoring protocols as described in the evalchemy \cite{evalchemy2023} setup.

\begin{table}[ht]
  \centering
  \scriptsize
  \setlength{\tabcolsep}{3pt} 
  \resizebox{\textwidth}{!}{%
    \begin{tabular}{@{} l r r r r r r r r @{}}
      \toprule
      Benchmark
        & \makecell{RekaAI\\Flash-3} 
        & \makecell{Nvidia\\LlaMa-3.1-Nemo-\\tron-Nano-8B-v1} 
        & \makecell{Qwen\\QWQ-32B} 
        & \makecell{OpenAI\\O1-mini} 
        & \makecell{LGAI\\ExaOne-32B} 
        & \makecell{Deepseek-R1\\Distill-Qwen-32B}
        & Deepseek-R1
        & \makecell{Apriel\\Nemotron-15b-\\thinker} \\
      \midrule
      GPQA Diamond       
        & 53.53 
        & 54.04
        & 66.67
        & 60
        & 65.15
        & 62.1
        & 71.5
        & 57.4  \\
      MATH-500
        & 84
        & 89.6
        & 90.8
        & 90
        & 91.6
        & 94.3
        & 97.5
        & 91.6  \\
      AIME’24
        & 38.67
        & 61.99
        & 81.33
        & 63.6
        & 76
        & 72.6
        & 79.8
        & 73.33 \\
      AIME’25
        & 25.33
        & 48.66
        & 68.67
        & 54.8
        & 64.67
        & 55.2
        & 66.8
        & 60    \\
      MMLU-Pro
        & 66.85
        & 61.53
        & 78.97
        & 80.3
        & 73.89
        & 77.2
        & 84
        & 73.42 \\
      AMC23
        & 83.1
        & 93.49
        & 98.5
        & 92.5
        & 95
        & 95.5
        & 99
        & 95    \\
      \makecell{LiveCodeBench (v5)\\}
        & 44.8
        & 53.2
        & 65.9
        & 53.8
        & 62.4
        & 57.2
        & 65.9
        & 54.56 \\
      \bottomrule
    \end{tabular}%
  }
  \vspace{4mm}
   \caption{Evaluation (pass@1) on academic research oriented and competitive benchmarks, including graduate-level question answering (GPQA Diamond), advanced mathematics problem solving (MATH-500, AIME’24, AIME’25, AMC23), multidisciplinary reasoning (MMLU-Pro), and code generation with functional correctness (LiveCodeBench v5).}
  \label{tab:benchmark-results}
  \vspace{1ex}
  \raggedright\footnotesize%
\end{table}

Our evaluations indicate that \textsc{Apriel-Nemotron}-15B-\textsc{Thinker} excels in advanced mathematical and logical reasoning and punches above its weight in general domain understanding. However, generating correct, executable Python remains a relative bottleneck. 

\subsection{Token Utilization}

\begin{figure}[h]
  \centering
  \begin{tikzpicture}
    \begin{axis}[
      width=0.9\linewidth,
      height=7.5cm,
      ybar=8pt,
      bar width=12pt,
      enlarge x limits=0.20,
      symbolic x coords={AIME-24,AIME-25,GPQA Diamond,MATH-500},
      xtick=data,
      xticklabel style={rotate=45,anchor=east},
      ylabel={Thinking Tokens Consumed},
      ymin=0,
      ytick={0,5000,10000,15000,20000},
      yticklabels={0,5\,k,10\,k,15\,k,20\,k},
      nodes near coords,
      nodes near coords style={
        font=\scriptsize,
        text=black
      },
      legend style={
        at={(0.5,1.05)},
        anchor=south,
        legend columns=3,
        /tikz/every even column/.append style={column sep=1cm}
      },
    ]
      \addplot+[draw=black, fill=green!60!black] coordinates {
        (AIME-24,8627)   (AIME-25,10332)
        (GPQA Diamond,5407) (MATH-500,2511)
      };
      \addplot+[draw=black, fill=gray!60] coordinates {
        (AIME-24,13422)  (AIME-25,16398)
        (GPQA Diamond,7575) (MATH-500,4437)
      };
      \addplot+[draw=black, fill=yellow!60!black] coordinates {
        (AIME-24,17528)  (AIME-25,19707)
        (GPQA Diamond,10568) (MATH-500,5317)
      };
      \legend{
        Apriel-Nemotron-15b-thinker,
        QWQ-32B,
        LG-ExaOne-32B
      }
    \end{axis}
  \end{tikzpicture}
  \caption{Thinking Token Consumption by Model Across Academic Reasoning Tasks}
  \label{fig:thinking-tokens}
\end{figure}
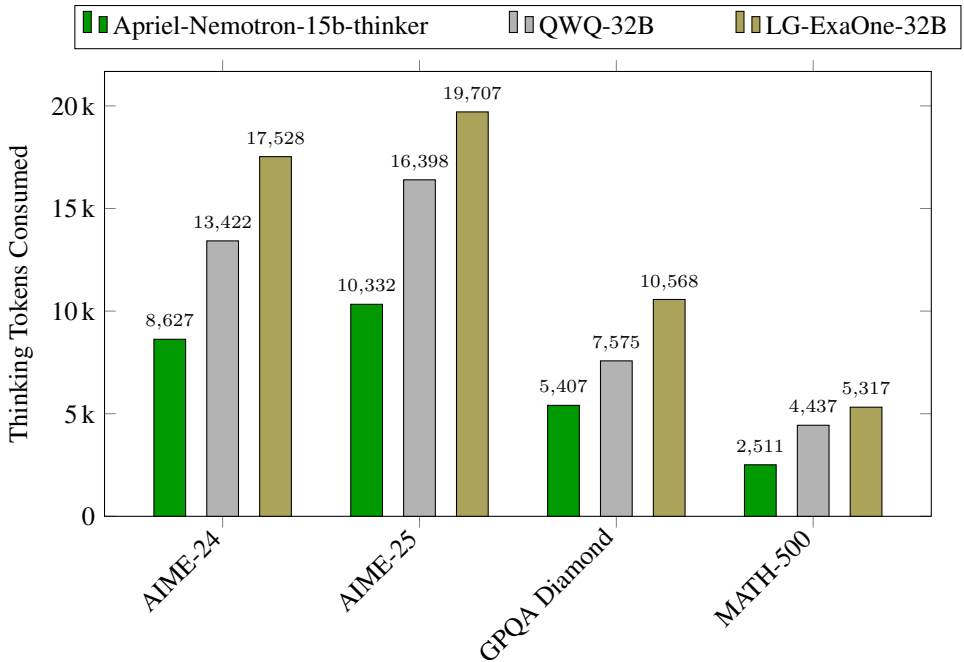

The comparison of the number of “thinking tokens” consumed by three models—Apriel-Nemotron-15b-thinker, QWQ-32B, and LG-ExaOne-32B—when solving four academic benchmarks: AIME-24, AIME-25, GPQA Diamond, and MATH-500. Across every task, the larger models (QWQ-32B and LG-ExaOne-32B) expend substantially more tokens than Apriel-Nemotron-15b-thinker. The LG-ExaOne-32B consistently uses the most tokens, peaking on AIME-25, while Apriel’s lightweight architecture remains the most efficient, especially on simpler arithmetic (MATH-500) and logic puzzles (GPQA Diamond). This visualization highlights the trade-off between computational “thinking” cost and model scale when tackling diverse mathematical and logic problems.

\begin{ack}
Use unnumbered first level headings for the acknowledgments. All acknowledgments
go at the end of the paper before the list of references. Moreover, you are required to declare
funding (financial activities supporting the submitted work) and competing interests (related financial activities outside the submitted work).
More information about this disclosure can be found at: \url{https://neurips.cc/Conferences/2024/PaperInformation/FundingDisclosure}.

Do {\bf not} include this section in the anonymized submission, only in the final paper. You can use the \texttt{ack} environment provided in the style file to automatically hide this section in the anonymized submission.
\end{ack}


\printbibliography
\newpage
\section{Contributions and Acknowledgments}
\label{sec:contributors}
\begin{tabular}{p{0.45\textwidth} p{0.45\textwidth}}
\textbf{Core Contributors} & \\[0.35em]
Shruthan Radhakrishna & Data \& Mid-Training \\[0.25em]
Soham Parikh & Mid-Training \& Architecture \\ [0.25em]
Gopal Sarda & Post-Training \\ [0.25em]
Anil Turkkan & Post-Training \\ [0.25em]
Quaizar Vohra & Post-Training \\ [0.25em]
Raymond Li & Post-Training \\[0.25em]
Dhruv Jhamb & Post-Training\\[0.25em]
Kelechi Ogueji & Post-Training\\[0.25em]
Aanjaneya Shukla & Benchmarking \\ [0.25em]
Oluwanifemi Bamgbose & Benchmarking \\ [0.35em]
\end{tabular}

\begin{tabular}{p{0.45\textwidth} p{0.45\textwidth}}
\textbf{Contributors}\\[0.3em]
Toby Liang\\[0.25em]
Luke Kumar\\[0.25em]
Oleksiy Ostapenko\\[0.25em]
Shiva Krishna Reddy Malay\\[0.25em]
Aman Tiwari\\[0.25em]
Tara Bogavelli\\[0.25em]
Vikas Yadav\\[0.25em]
Jash Mehta\\[0.25em]
Saloni Mittal\\[0.25em]
Akshay Kalkunte\\[0.25em]
Pulkit Pattnaik\\[0.25em]
Khalil Slimi\\[0.25em]
Anirudh Sreeram\\[0.25em]
Jishnu Nair\\[0.25em]
Akintunde Oladipo\\[0.25em]
Shashank Maiya\\[0.25em]
Khyati Mahajan\\[0.25em]
Rishabh Maheshwary\\[0.25em]
Masoud Hashemi\\[0.25em]
Sai Rajeswar Mudumba\\[0.35em]
\end{tabular}


\begin{tabular}{p{0.45\textwidth} p{0.45\textwidth}}
\textbf{Leads \& Management}\\[0.3em]
Sathwik Tejaswi Madhusudhan & Overall Technical Lead\\[0.25em]
Torsten Scholak & Pre-Training and Architecture Lead\\[0.25em]
Sebastien Paquet & Research Manager\\[0.25em]
Sagar Davasam & Applied Research Manager\\[0.25em]
Srinivas Sunkara & VP, Applied AI Research\\[0.5em]
\end{tabular}\\[0.3em]

\textbf{Acknowledgment}\\[0.4em]
We acknowledge Anil Madamala for his leadership in evaluations and benchmarking, and Segan Subramanian and Vipul Mittal for their leadership in data infrastructure.\\[0.25em]
We thank researchers at Nvidia for sharing detailed insights and data from their work in building reasoners. This greatly accelerated our research and we recognize the same with our model naming convention.

\end{document}